\documentclass{article}
\usepackage{ijcai26}

\usepackage{times}

\usepackage{soul}
\usepackage{url}
\usepackage[hidelinks]{hyperref}
\usepackage[utf8]{inputenc}
\usepackage[small]{caption}
\usepackage{graphicx}
\usepackage{amsmath}
\usepackage{amsthm}
\usepackage{booktabs}
\usepackage{algorithm}
\usepackage{algorithmic}
\usepackage[switch]{lineno}

\usepackage{amsmath,amssymb,amsfonts}
\usepackage{diagbox}
\usepackage{textcomp}
\usepackage{xcolor}
\usepackage{bm}
\usepackage{enumitem}
\usepackage{array,multirow}
\usepackage{siunitx}
\usepackage{xfp} 
\usepackage{subcaption}
\usepackage{makecell}
\usepackage[normalem]{ulem}
\usepackage{xparse}

\newtheoremstyle{myplain}
  {\topsep}{\topsep}
  {\itshape}{}
  {\bfseries}{}   
  {5pt plus 1pt minus 1pt}
  {\thmname{#1}\thmnumber{ #2}\thmnote{ (#3)}}
\theoremstyle{myplain}
\newtheorem{assumption}{Assumption}[section]

\newtheorem{definition}{\bf  Definition}

\NewDocumentCommand{\fkm}{m o}{%
  \IfNoValueTF{#2}
    {\textcolor{red}{#1}}%
    {\sout{#1}\,\textcolor{red}{#2}}%
}

\title{Invariant Graph Representations for Continuous-Time Dynamic Graphs Under Distribution Shifts}

\author{
Lanting Fang$^1$ \and
  Yulian Yang$^2$ \and
    Yawei Zhang$^1$\and 
  Shanshan Feng$^3$\thanks{Shanshan Feng is the corresponding author.} \and
  Kaiyu Feng$^1$\And
  Hanning Yuan$^1$ \\
    \affiliations
$^1$Beijing Institue of Technology\\
$^2$ 	Southeast University \\
$^3$	Wuhan University\\
    \emails
ltfang@bit.edu.cn, 
406674524@qq.com,
z15334025555@163.com,
\{fengky, yhn6\}@bit.edu.cn
}

\begin{document}

\maketitle

\begin{abstract}
Continuous-Time Dynamic Graphs (CTDGs) enable fine-grained modeling of evolving relational systems.  However, most existing CTDG representation learning methods are tailored to in-distribution settings and exhibit limited robustness under out-of-distribution (OOD) shifts. Although recent causal approaches  learn invariant representations via interventions, they are primarily designed for static or discrete-time graphs and become computationally prohibitive for CTDGs due to the combinatorial explosion of structural and temporal variations.
To address these challenges, we propose CIR, a framework grounded in a novel structural causal model termed the ICCM. To avoid exhaustive interventions, we leverage the Normalized Weighted Geometric Mean (NWGM) to efficiently approximate interventional predictions. We further instantiate ICCM within a practical deep learning architecture that jointly captures invariant structural and temporal patterns through dedicated subgraph extractors, and maintains an environment memory bank to model distributional shifts across evolving contexts. Extensive experiments demonstrate that CIR consistently outperforms existing methods under diverse OOD scenarios.
\end{abstract}

\section{Introduction}
Many real-world systems exhibit continuously evolving relational structures, which are naturally modeled as dynamic graphs \cite{zhang2025dynamic}. Existing dynamic graph models can be broadly categorized into Discrete-Time Dynamic Graphs (DTDGs) \cite{DTDG2,DTDG3}, which represent evolution as a sequence of temporal snapshots, and Continuous-Time Dynamic Graphs (CTDGs) \cite{CTDG2025_1,CTDG2025_2,CTDG1,CTDG3}, which capture fine-grained structural changes via continuous node and edge updates. 
Although existing methods have achieved significant success in modeling temporal dynamics, they predominantly overlook robustness under  out-of-distribution (OOD)  scenarios.

\begin{figure}[h]
    \centering   \includegraphics[width=2.2in]{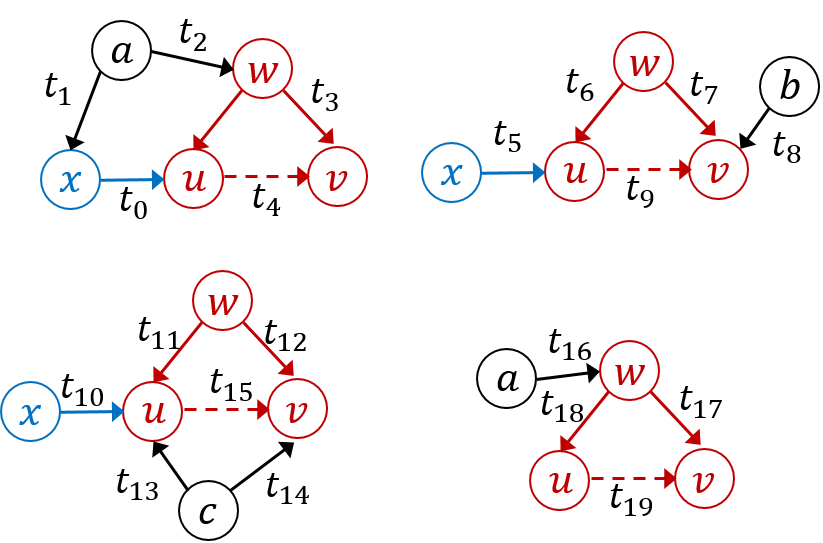}
    \caption{Examples of training data.}
    \label{fig:shortcut}
\end{figure}

Therefore, this paper focuses on invariant representation learning for CTDGs under OOD scenarios.
While techniques such as subgraph extraction \cite{SUNNYGNN,AdaDyGNN} and attention mechanisms \cite{ranjan2020asap} can partially alleviate distribution shifts, they remain vulnerable to exploiting spurious correlations \cite{chang2020invariant}.
For example, in the link prediction setting illustrated in Figure~\ref{fig:shortcut}, a triadic closure pattern (highlighted in red) constitutes the invariant subgraph underlying the formation of the target link $(u, v)$. However, because a bridging link $(x, u)$ (highlighted in blue), also appears before $(u, v)$ in most training graphs, the model may incorrectly treat $(x, u)$ as the invariant subgraph for $(u, v)$. As a result, at test time, the model may falsely predict the existence of $(u, v)$ based solely on $(x, u)$, ignoring the absent of the triadic closure pattern.

Recent structural causal model (SCM) based methods \cite{casual1,CAL,causal2,DisC} have been proposed to mitigate spurious correlations by identifying and intervening on invariant subgraphs, either across multiple environments or through sampling-based interventions. However, these approaches implicitly assume a fixed or bounded graph structure, where the set of nodes and candidate edges remains stable over time, as in static graphs or DTDGs. Under this assumption, causal invariance can be meaningfully defined and estimated over a shared structural support. In contrast, CTDGs allow nodes and edges to be added or removed at arbitrary time points, resulting in a non-stationary and exponentially expanding space of graph topologies. This  fundamentally challenges existing SCM-based intervention mechanisms.


In this paper, we propose  \emph{\underline{C}ontinuous-time \underline{I}nvariant  \underline{R}epresentation (CIR)}, a novel method that captures invariant representation and mitigates OOD effects from a new SCM perspective.
To effectively address OOD generalization in CTDGs, 
CIR is designed to satisfy three essential properties: (1) robust mitigation of OOD distribution shifts, which is fundamental for generalization across unseen environments; (2) simultaneous capture of invariant subgraphs across both structural and temporal dimensions, which is crucial in CTDGs where graph topology and event timing evolve continuously; and (3) computationally efficient causal interventions, necessitated by the exponentially large space of possible structural–temporal subgraphs.

CIR is theoretically grounded in a structural causal model termed the Independent and Confounded Causal Model (ICCM), which explicitly models both in-distribution and OOD environments. Directly intervening on all possible structural and temporal subgraphs is computationally infeasible. To address this challenge, we leverage the Normalized Weighted Geometric Mean (NWGM) \cite{NWGM} to efficiently approximate the expected interventional outcome, substantially reducing computational cost.
We instantiate ICCM within a practical deep learning framework. The architecture first employs dual subgraph extractors to capture causal structural and temporal patterns. An environment memory bank is then maintained to learn environment-specific representations, which are used to condition the prediction process. To approximate the expected outcome of causal interventions defined in ICCM, we aggregate environment-conditioned predictions using the NWGM. Finally, the causal structural and temporal patterns are encoded into invariant representations, and the model performs predictions based on the learned invariant representations together with the environment representations.
The main contributions of this paper are summarized as follows:
\begin{itemize}[leftmargin=0.5cm]
\item To the best of our knowledge, we are the first to investigate invariant representation learning problem for CTDGs under OOD shifts. 
\item We introduce ICCM, a novel structural causal model that characterizes invariant causal subgraphs in CTDGs, and develop an efficient intervention strategy based on the NWGM to approximate interventional predictions, avoiding exhaustive enumeration of structural and temporal subgraphs.
\item We design a framework named CIR that instantialize ICCM, enabling the joint learning of invariant structural and temporal representations along with environment-specific features.
\end{itemize}

\section{Related Work}
\label{sec:relatedworkd}
In this section, we review two areas closely related to our study: dynamic graph neural networks and invariant graph representation.

\paragraph{Dynamic Graph Neural Networks.}
Dynamic graph neural networks encompass two primary classifications: Discrete-Time Dynamic Graphs (DTDGs)  \cite{DTDG3,SEIGN} and Continuous-Time Dynamic Graphs (CTDGs) \cite{CTDG1,CTDG3,li2023zebra}. DTDGs comprise a sequence of static graph snapshots captured at regular time intervals \cite{DySAT,Netwalk,dynnode2vec}. 
CTDGs capture the evolution of graphs by considering modifications on the graph that occur continuously rather than discretely at predefined time steps \cite{TGAT,Dyrep,OTGNet}. 
However, these GNNs focus primarily on modeling graph dynamics while overlooking robustness to OOD scenarios.

\paragraph{Invariant Graph Representation.}
Invariant learning enhances the robustness of GNNs by identifying stable, environment-invariant patterns. Existing approaches fall broadly into two categories: disentangled learning and causal learning.
Disentangled learning \cite{yuan2025structure,ning2025summary,OOD-GCL} aims to learn representations in which latent factors are separated and interpretable, capturing distinct variations in the data. Such representations have been shown to improve robustness to adversarial attacks and enhance generalization. However, these methods are limited in their ability to identify causal variables that remain invariant in the presence of shortcuts or spurious correlations.
Causal inference methods \cite{causal3,DIR} focus on unveiling and comprehending the true causal variables that drive observed phenomena. On real-world graphs, uncovering these causal variables becomes an act of explanation, revealing the ``why'' behind intricate relationships.
However, most of these methods are largely restricted to static settings. 
To extend causal inference into dynamic scenarios, DIDA \cite{DIDA} has been proposed as an invariant rational discovery framework for DTDGs. However, DIDA requires constructing an intervention set for  each snapshot, making it computationally intensive when applied to CTDGs. EAGLE \cite{EAGLE} employs a VAE-based model as an environment sample generation network. However, when the number of environments is large and the available sample size is limited, the generated samples may not adequately represent the underlying environments, thereby impairing generalization performance.
In this work, we focus on the more realistic and challenging setting of CTDGs, where interactions evolve continuously rather than at discrete time intervals. Existing causal inference methods are generally inapplicable to this setting because they predominantly rely on snapshot-based assumptions, which fail to capture the fine-grained temporal dynamics inherent in CTDGs.

\section{Problem Formulation}
\label{sec:formulation}

\begin{definition}[CTDG]
A Continuous Time Dynamic Graph (CTDG) $G = (\mathcal{V}, \mathcal{E}, \mathcal{T})$ consists of a set of vertices $\mathcal{V}$, a set of edges $\mathcal{E}$, and a continuous time domain $\mathcal{T}$. The graph evolves over $t\in \mathcal{T}$ through edge additions, removals, or changing characteristics. We represent $G$ as a sequence of time-stamped edges $G = \langle e_{ij}(t_k)\rangle$, where $e_{ij}(t_k)$ denotes an interaction between node $v_i$ and $v_j$ at time $t_k$ with feature vector $x_{ij}^e(t_k)$. Each node $v_i$ has a feature vector $x_i^n$. 
\end{definition}

Given a CTDG $G = (\mathcal{V}, \mathcal{E})$, the task is to predict whether an interaction $e_{ij}(t)$ will occur at a future time $t$, based on the historical sequence of nodes, edges, and their features. To evaluate OOD robustness, the model is trained on one environment and tested on CTDGs from other environments, where structural and temporal distributions may differ.



\section{Independent and Confounded Causal Model}
\label{sec:casual_look}
In this section, we introduce a novel structural causal model framework, termed Independent and Confounded Causal Model (ICCM).

In dynamic graphs, the existence of an edge can be influenced by both the structural topology (e.g., triadic closure) and temporal dynamics (e.g., temporal burst)~\cite{GraphMixer}.
These mechanisms may act independently or become entangled through latent confounders, and their influence can vary  across environments. Existing causal formulations typically conflate structural and temporal effects, limiting their ability to disentangle spurious correlations under distribution shifts.

To explicitly capture both independent causal effects and environment-dependent confounding, ICCM models structural and temporal causal processes as distinct but potentially confounded components.
We formalize ICCM by inspecting the causalities among nine variables: the structural causal subgraph $C_s\subseteq G$, the temporal causal subgraph $C_t\subseteq G$,  the hidden representations  $H^T$ and $H^S$, the structural  confounder $\mathcal{D}_s$, the temporal confounder $\mathcal{D}_t$, the edge existence probability in the in-distribution environment $Y^I$, the edge existence probability in OOD environments $Y^O$, and the final prediction result $Y$.
Assumption~\ref{ass:ICCM} summarizes the core principles underlying ICCM:
\begin{assumption}[\textbf{ICCM}]
\label{ass:ICCM} 
\begin{align*}
 &H^S=  f_{s}(C_s), \quad  H^T=  f_{t}(C_t),  \quad Y^I =  f^{I}_o(H^S, H^T) \\
&Y^C=  f^C_o(H^S,H^T, \mathcal{D}_s, \mathcal{D}_t), 
\quad \text{s.t.} \quad C_s \bot \mathcal{D}_s, \quad C_t \bot \mathcal{D}_t, \\
& Y = f_o(Y^I, Y^C).
 \end{align*}
\end{assumption}
\noindent In this assumption, $C_s$ and $C_t$ are  causal   structural and temporal subgraph.  These subgraphs are encoded into latent representations $H^S$ and $H^T$ via $f_s(\cdot)$ and $f_t(\cdot)$. 
The in-distribution prediction $Y^I$ is determined by the outcome function $f_o^I(H^S, H^T)$:
\begin{equation}
\label{eq:overview_ICM}
Y^I  = f_o^I(H^S, H^T)= \sigma \bigl( W^I(W^I_1H^S+W^I_2H^T )\bigr).
\end{equation}

In OOD environments, however, structural and temporal confounders introduce spurious correlations that degrade model generalization. Specifically, structural confounders establish backdoor paths  $C_s\leftarrow C_b^s \rightarrow Y$ and $C_s \leftarrow U_s \rightarrow Y$, where $C_b^s=G\setminus C_s$, and $U_s$ represent unobserved variables.
These paths induce spurious correlations between $C_s$ and  $Y^C$.
Temporal confounders exert a similar influence. To mitigate these spurious correlations, we employ the \emph{do-operation} introduced by Pearl~\cite{pearl2016causal}. The \emph{do-operation} allows us to block backdoor paths, thereby eliminating spurious correlations and isolating the invariant causal effect. Using the \emph{do-operation}, the prediction
$Y^C$ can be formulated as:
\begin{equation}
\label{eq:adjust_s}
\begin{aligned}
Y^C=&  f^C_o(H^S,H^T, \mathcal{D}_s, \mathcal{D}_t) \quad \text{s.t.} \quad C_s \bot \mathcal{D}_s, \quad C_t \bot \mathcal{D}_t,\\
=&\mathbb{E}_{u\in \mathcal{D}_S, v\in \mathcal{D}_T}\left[f_o^C(H^S, H^T, u, v) \right] \\
\end{aligned}
\end{equation}
where $\mathcal{D}_s=\{ C^s_b, U_s\}$ and $\mathcal{D}_t=\{C^t_b, U_t\}$, respectively,  $f^C_o$ is output network network:
\begin{equation}
\label{eq:overview_CCM}
\begin{aligned}
f^C_o(H^S,H^T,u, v)=   &  \sigma \bigl(W^c_1f_y^s(H^S)+W^c_2f_y^t(H^T) \\ \qquad\qquad \qquad   \qquad  &+W^c_3f_y^u(u)
 +W^c_4f_y^v(v) \bigr).
 \end{aligned}
\end{equation}
where 
$W^*_*$ denotes the model parameters, $f_y^*(\cdot)$ denotes linear network, $\sigma$ denotes the activation function. 
However, we observe
Equation \ref{eq:adjust_s} is computationally intractable. It necessitates enumerating all possible pairs of invariant causal subgraphs across the space of structural confounders $\mathcal{D}_s$ and  temporal confounders $\mathcal{D}_t$.
In the context of large-scale CTDGs, this combinatorial explosion of potential subgraph configurations leads to prohibitive computational costs, making exhaustive intervention infeasible.

To address this issue,  we propose an efficient intervention process by leveraging the NWGM approximation \cite{NWGM}. The NWGM provides a computationally efficient way to estimate the expectation of a softmax or sigmoid output by moving the expectation inside the non-linearity:
\begin{equation}
      \mathbb{E}_{d\sim \mathcal{D}}[\sigma\bigl(f(x)\bigl)] \approx  \sigma\bigl(\mathbb{E}_{d\sim \mathcal{D}}[f(x)]\bigl).
\end{equation}
By applying the NWGM approximation, the interventional prediction in Equation \ref{eq:adjust_s} can be reformulated as:
\begin{equation}
\label{eq:overview_s}
\begin{aligned}
Y^C \approx & 
\sigma \bigl( \mathbb{E}_{u\in \mathcal{D}_s, v\in  \mathcal{D}_t}[ W^c_1f_y^s(H^S)+W^c_2f_y^t(H^T)  \\
&\qquad\qquad \qquad  +W^c_3f_y^u(u)  +W^c_4f_y^v(v) ]\bigr), \\
=  & \sigma \bigl(W^c_1f_y^s(H^S)+W^c_2f_y^t(H^T) +W^c_3\mathbb{E}_{u\sim \mathcal{D}_s}[f_y^u(u)]\\
&\qquad\qquad \qquad+W^c_4\mathbb{E}_{v\sim \mathcal{D}_t}[f_y^v(v)]\bigr).  
\end{aligned}
\end{equation}
Note that $\mathcal{D}_s$ and $\mathcal{D}_t$ denote abstract latent confounders, which may themselves correspond to complex subgraph patterns. While the confounder space can be large, Equation~\ref{eq:overview_s} does not require explicit enumeration of all possible confounders. Instead, it only requires expectation over a representative set of confounding modes, which can be approximated via confounder prototypes (Explained in Section~\ref{sec:network:confounder}.

\section{Neural Instantiation of ICCM}
\label{sec:Implementation}
\begin{figure*}[t]
    \centering
\includegraphics[width=.9\linewidth]{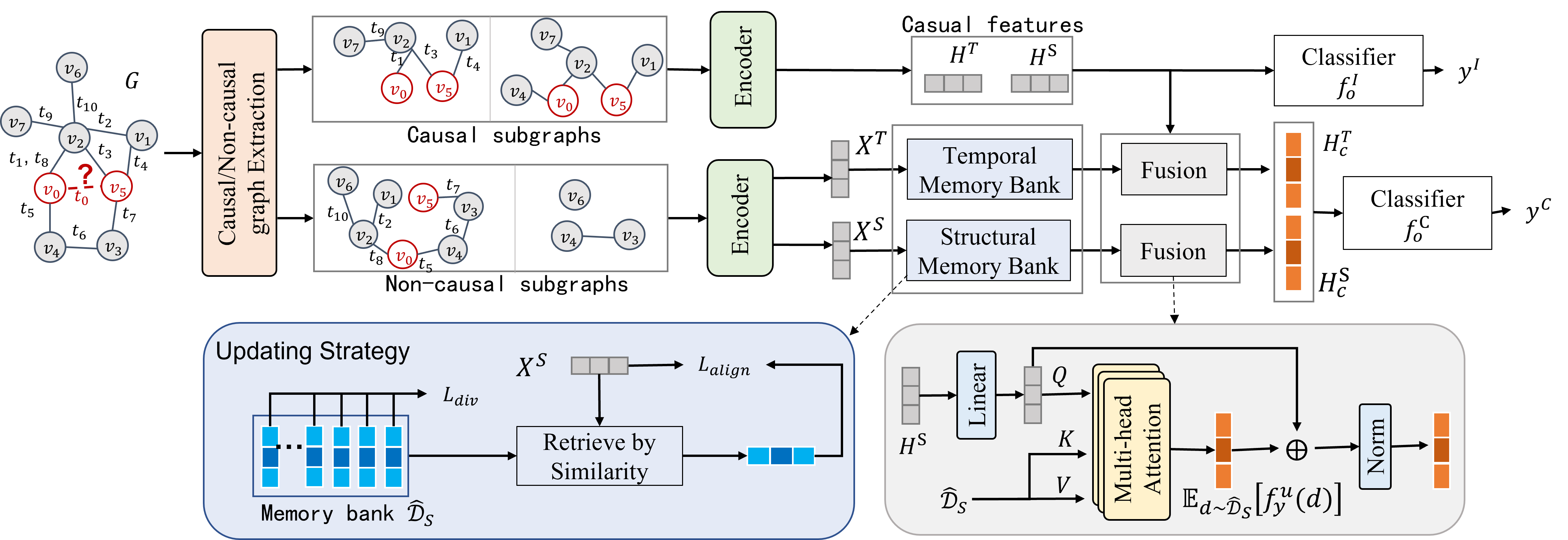}
    \caption{The deep learning implementation of CIR.}
    \label{fig:framework}
\end{figure*}

In this section, we describe how the proposed ICCM is instantiated within a neural architecture by mapping latent causal variables to learnable components. The complete architecture is illustrated in Figure~\ref{fig:framework}.

While the theoretical ICCM assumes access to causal subgraphs $C_s$ and $C_t$, these variables are typically latent and unobserved in real-world scenarios. To address this, we employ causal subgraph extractors to infer structural and temporal subgraphs directly from the input data. 
 The extractors identify invariant structural and temporal patterns and produce hidden representations $H^S$ and $H^T$. 
 
 For the in-distribution perspective, these representations are passed to the outcome function $f_o^I$ to predict the label $y^I$. For the out-of-distribution (OOD) perspective, as described in Equation~\ref{eq:overview_s}, the interventional prediction requires the expected representations of structural and temporal confounders:
$\mathbb{E}_{u \sim \mathcal{D}_S}[f_y^u(u)] \quad \text{and} \quad \mathbb{E}_{v \sim \mathcal{D}_T}[f_y^v(v)]$.

To estimate these expectations without exhaustive enumeration, we maintain structural and temporal memory banks that approximate the latent confounder distributions $\mathcal{D}_s$ and $\mathcal{D}_t$. A multi-head attention-based fusion module then aggregates the memory entries to compute the expected environmental context, which approximates the expectation over latent confounder effects.
Finally, the invariant representations $H^S$ and $H^T$ are integrated with the computed expectations from the memory banks and fed into the outcome function $f_o^C$ to produce the OOD prediction $y^C$.
We term this framework Continuous Invariant Representation (CIR). 

\subsection{Extracting and Encoding Causal/Non-causal Subgraphs}
\paragraph{Temporal extraction and encoding.}
Given a dynamic graph $G$ and a target node pair $(u, v)$ for prediction, our goal is to identify historical temporal interactions that causally contribute to the potential existence of an edge between $u$ and $v$. Unlike heuristic time-window approaches, we aim to extract temporally relevant evidence that is both \textbf{time-aware} and \textbf{pair-specific}, aligning with the causal assumptions in ICCM.
we first extract the temporal edges associated with nodes $u$ and $v$ denoted by $S_G^u$ and $S_G^v$, respectively.
Each edge $e_{ui}(t_k) \in S_G^u$ is encoded using a continuous-time representation following \cite{GraphMixer}:
$\cos \bigl((t_0-t_k) \omega \bigr)$, 
 where $t_0$ is the prediction timestamp, $\omega=\left\{\alpha^{-(i-1)/ \beta}\right\}_{i=1}^d$ defines a set of multi-scale temporal frequencies, with $\alpha$ and $\beta$ representing hyperparameters. This encoding is combined with its corresponding edge features as $[\cos \bigl(\left(t_0-t_k\right) \omega\bigr) \| \mathbf{x}_{ui}^{e}\left(t_k\right)]$. 
 
A 1-layer MLP-mixer \cite{tolstikhin2021mlpmixer} is employed to encode the edge features,
\begin{equation}
    F_u=\text{MLPmixer}(\text{stack}(S_G^u)),
\end{equation}
where stack$(S_G^u)$  indicates the stacking of the edge features in sequence $S_G^u$. The MLP-Mixer enables global mixing across temporal edges, allowing the representation of higher-order temporal dependencies that are crucial for causal reasoning.

Intuitively, a historical interaction of $v$ is considered temporally causal for predicting $(u,v)$ only if it helps explain $u$’s behavior, and vice versa. Therefore, we introduce a cross-node attention mechanism.
The subgraph is  generated by:
\begin{equation}
\begin{aligned}
 & q_u = W^{m}_1 \text{Mean}( F_u), \quad K_v =W^{m}_2 (F_v) \\
 &q_v = W^{m}_1 \text{Mean}( F_v), \quad K_u =W^{m}_2 (F_u) \\
     &M^e_{v} = \text{Softmax}\left(\frac{q_u^T K_v}{\sqrt{d}}\right), \quad 
M^e_{u} = \text{Softmax}\left(\frac{q_v^T K_u}{\sqrt{d}}\right).
\end{aligned}    
\end{equation}
Here, $d$ denotes a specific hyperparameter, and $M^e_*[k]$ represents the importance score assigned to the $k$-th edge within $S_*$. This attention mechanism filters temporal evidence based on mutual relevance, rather than local frequency or recency alone.
Consequently, 
the highest top-$k$ scores in $M^e_v$ and $M^e_u$ are selected to construct the temporal causal subgraphs  $C_t$, while the remaining edges form the temporal non-causal subgraphs $\hat{B}_t$.
Finally, the temporal representations of the causal and non-causal graphs are encoded by:
\begin{equation}
\begin{aligned}
  & h^t_u=\text{MLPmixer}(\text{stack}(S_{C_t}^u), \quad
h^t_v = \text{MLPmixer}(\text{stack}(S_{C_t}^v), \\
&  x^t_u=\text{MLPmixer}(\text{stack}(S_{\hat{B}_t}^u), \quad
x^t_v = \text{MLPmixer}(\text{stack}(S_{\hat{B}_t}^v), 
\end{aligned}
\end{equation}
The final representations are obtained by concatenation: 
\begin{equation}
    H^T = [h^t_u || h^t_v], \quad X^T = [x^t_u || x^t_v].
\end{equation}
Causal temporal evidence is disentangled from confounding effects, supporting intervention-based learning.

\paragraph{Structural extraction and encoding.}
Our objective is to identify structural neighbors that causally support the interaction between $u$ and $v$, while filtering out background nodes that may introduce spurious correlations.
We first encode each node based on its  n-hop neighborhood within a recent time window $]t_0-T, t_0]$: $z_u=x_u^{n}+\operatorname{Mean}
\bigl(\{x_v^n \mid v \in \mathcal{N}^n(u ; t_0-T, t_0)\}\bigr)$. 
Here, $\mathcal{N}^n(u ; t_0-T, t_0)$ denotes the n-hop neighbors of node $u$ with valid temporal edges, where $T$ controls the structural temporal scope. 
The node mask matrices are computed  through the equations:
\begin{equation}
    M^n_{v} = \text{Softmax}\left(\frac{z_u^T Z_v}{\sqrt{d}}\right),  \quad M^n_{u} = \text{Softmax}\left(\frac{z_v^T Z_u}{\sqrt{d}}\right).
\end{equation}
Here, $Z_u$ and $Z_v$ is the stack of the encoded node features of all nodes in $\mathcal{N}^n(u ; t_0-T, t_0)$ and $\mathcal{N}^n(v ; t_0-T, t_0)$, respectively. The nodes with the highest top-$k$ scores in $M^n_{u}$ and $M^n_{v}$ are chosen to form the structural causal subgraph,  while the remaining edges form the temporal non-causal subgraphs $\hat{B}_s$.
The final structural representation $H^S$ is computed by:  
\begin{equation}
\begin{aligned}
    & h^s_u =x^{n}_u + \operatorname{Mean} \bigl(\{x^{n}_i| i\in \mathcal{N}^n_S(u) \}\bigr), \\ &h^s_v =x^{n}_v+ \operatorname{Mean}\bigl(\{x^{n}_i| i \in \mathcal{N}^n_S(v) \}\bigr), \\ &H^S = [h^s_u || h^s_v],   
\end{aligned}
\end{equation}
where $x^{n}_v$ represents the  node feature of $v$, $\mathcal{N}^n_S(u)$ represents the n-hop neighbors of $u$ in $C_s$.
The final structural representation $X^S$ for the non-causal subgraph $\hat{B}_s$ is obtained using the same encoding method applied to $C_s$.

\subsection{Memory Banks}\label{sec:network:confounder}
The memory bank is designed to capture all confounders, including both non-causal subgraphs and unobserved factors. 
Since unobserved factors are not directly observable, we approximate them using the representations of observable non-causal subgraphs, based on the insight that unobserved influences are often implicitly reflected in the structure of these subgraphs. 

As the dynamic graph evolves, the number of non-causal subgraphs can grow exponentially, making it infeasible to consider each instance individually.
To address this, each entry in the memory bank approximates a cluster of non-causal subgraphs, capturing the shared influence of similar confounders.
Specifically, the memory bank consists of $K$ prototype vectors stored in the  $\hat{\mathcal{D}}_* \in \mathbb{R}^{K \times d}$, where $K$ denotes the number of prototypes and $d$ represents the feature dimensionality. Each element of $\hat{\mathcal{D}}_*$ is initialized by sampling from a standard normal distribution, after which each prototype is normalized to ensure that they lie on a unit hypersphere.
The memory bank is updated during training via gradient descent, guided by two self-supervised loss functions: the \emph{alignment loss} and the \emph{diversity loss}. 

 Given non-causal graphs $X^T$ and $X^S$,
 the alignment loss is defined as:
\begin{equation} 
\begin{aligned}
       & z_T = \text{linear}(X^T)  \quad z_S = \text{linear}(X^S)\\
      & k_T = \arg\max_k \,  z_T \hat{\mathcal{D}}_T^\top[k] \quad k_S = \arg\max_k \,  z_S \hat{\mathcal{D}}_S^\top[k] \\
&\mathcal{L}_{\text{align}} = 1 -  (z_T \hat{\mathcal{D}}_T[k_T]+z_S \hat{\mathcal{D}}_S[k_S]), 
\end{aligned}
\end{equation}
Here, $z_T$ and $z_S$ are encoded features of $X^T$ and $X^S$,  respectively. The term $z_* \hat{\mathcal{D}}_*^\top[k_*]$ (with $*$ representing either $T$ or $S$) measures the similarity between the encoded feature $z_*$  and the 
$k_*$-th prototype in $\hat{\mathcal{D}}_*$.
Each sample is assigned to the prototype with the highest similarity. When $z_*$ is well aligned with $\hat{\mathcal{D}}_*[k]$, the loss contribution for that sample is minimal.

To prevent redundancy among the prototypes, we impose a diversity loss:
\begin{equation}
    \mathcal{L}_{\text{div}} = \frac{1}{K} (\sum_{i=1}^{K}\sum_{j=1}^{K} \hat{\mathcal{D}}_S[i] \cdot \hat{\mathcal{D}}_S[j]+\sum_{i=1}^{K}\sum_{j=1}^{K} \hat{\mathcal{D}}_T[i] \cdot \hat{\mathcal{D}}_T[j]).
\end{equation}
Minimizing this loss encourages the prototypes to be as orthogonal as possible, ensuring that each captures a distinct cluster of non-causal subgraphs.

The overall regularization loss is the sum of the alignment and diversity losses:
\begin{equation}
    \mathcal{L}_{\text{reg}} = \mathcal{L}_{\text{align}} + \mathcal{L}_{\text{div}}.
\end{equation}
This loss is backpropagated during training to update both the encoder parameters and the memory bank, thereby refining the prototypes to more accurately represent distinct clusters of non-causal subgraphs.

\subsection{Fusion}
As shown in Equations~\ref{eq:overview_s}, the fusion module combines the causal graph representations ($H^S$ and $H^T$) with the expected contributions of confounders, as approximated by the memory banks ($\hat{\mathcal{D}}_S$ and $\hat{\mathcal{D}}_T$). This step implements the ICCM-inspired interventional reasoning, where the model accounts for potential spurious effects from non-causal subgraphs and unobserved factors.
We approximate the expected effect of confounders as:
\begin{equation}
\begin{aligned}
&Q_S = \text{Linear}(H^S) \quad  K_S=\hat{\mathcal{D}}_S\quad  V_S= \hat{\mathcal{D}}_S \\
       &\mathbb{E}_{u \sim \hat{\mathcal{D}}_S}\left[f_y^u(u)\right] = \operatorname{Multi-head}(Q_S, K_S, V_S), \\
 &Q_T = \text{Linear}(H^T) \quad  K_T=\hat{\mathcal{D}}_T\quad  V_T= \hat{\mathcal{D}}_T \\
       &\mathbb{E}_{v \sim \hat{\mathcal{D}}_T}\left[f_y^v(v)\right] = \operatorname{Multi-head}(Q_T, K_T, V_T).
\end{aligned}
\end{equation}

The attention output is then fused with the original query using a residual connection, followed by dropout and layer normalization:
\begin{equation}
\begin{aligned}
& H^S_c =\text{LayerNorm}\bigl(Q+Dropout(  \mathbb{E}_{u \sim \hat{\mathcal{D}}_S}\left[f_y^u(u)\right] )\bigr) \\
&H^T_c =\text{LayerNorm}\bigl(Q+Dropout(  \mathbb{E}_{v \sim \hat{\mathcal{D}}_T}\left[f_y^v(v)\right] )\bigr).
\end{aligned}
\end{equation}
The fused features $H^S_c$  and $H^T_c$ are subsequently processed by a feedforward network and output the final prediction $y^C$.

\subsection{Optimization}
We define the overall learning objective  as:
\begin{equation}
\begin{aligned}
         \mathcal{R}_i &= \frac{1}{|\mathcal{E}|} \sum_{e \in \mathcal{E}} \Bigl[ y_e \log(y_e^I) + (1-y_e) \log(1-y_e^I) \Bigr],\\
          \mathcal{R}_c &= \frac{1}{|\mathcal{E}|} \sum_{e \in \mathcal{E}} \left[ y_e \log(y_e^C) + (1-y_e) \log(1-y_e^C) \right],\\
         \mathcal{L} &= \lambda_i \, \mathcal{R}_i(y^I, y) +  \lambda_c \, \mathcal{R}_c(y^C, y) + \lambda_r \, \mathcal{L}_{\text{reg}},
\end{aligned} 
\end{equation}
where $\lambda_i$, $\lambda_c$, and $\lambda_r$ are hyperparameters,
$\mathcal{E}$ denotes the training dataset, which comprises pairs of positive and negative samples. Positive samples are derived from the original edge sets, while negative samples are generated by replacing the destination nodes with randomly sampled nodes from the vocabulary, ensuring a balanced ratio. The variable $y_e$ indicates the ground-truth label for edge $e$, with $y_e = 1$ for positive samples and $y_e = 0$ for negative samples.

\begin{table*}[t]
\centering
\resizebox{\linewidth}{!}{
\begin{tabular}{c|ccc|ccc|ccc|ccc}
\hline
\textbf{Model} & \multicolumn{3}{c}{\textbf{MOOC}} & \multicolumn{3}{c}{\textbf{REDDIT}} & \multicolumn{3}{c}{\textbf{WIKI}} & \multicolumn{3}{c}{\textbf{UCI}} \\ \hline
Ratio & 0.2 & 0.4 & 0.6 & 0.2 & 0.4 & 0.6 & 0.2 & 0.4 & 0.6 & 0.2 & 0.4 & 0.6 \\ \hline
TGN & 90.39{\scriptsize ±1.27} & 65.49{\scriptsize ±0.81} & 61.99{\scriptsize ±2.21} & 95.44{\scriptsize ±1.09} & 87.70{\scriptsize ±3.55} & 72.69{\scriptsize ±0.31} & 68.56{\scriptsize ±2.03} & 61.92{\scriptsize ±2.63} & 64.97{\scriptsize ±0.65} & 68.93{\scriptsize ±1.41} & 68.43{\scriptsize ±3.58} & 64.68{\scriptsize ±2.47} \\
TGAT & 93.75{\scriptsize ±0.49} & 85.25{\scriptsize ±1.21} & 73.08{\scriptsize ±1.49} & 97.32{\scriptsize ±0.14} & 95.34{\scriptsize ±0.28} & 89.21{\scriptsize ±0.48} & 90.90{\scriptsize ±0.78} & 74.72{\scriptsize ±0.74} & 73.07{\scriptsize ±2.74} & 77.66{\scriptsize ±1.17} & 74.55{\scriptsize ±0.46} & 72.59{\scriptsize ±0.54} \\
GraphMixer & \textbf{98.06{\scriptsize ±0.27}} & 88.11{\scriptsize ±4.62} & 48.71{\scriptsize ±3.02} & \textbf{98.41{\scriptsize ±0.33}} & 78.85{\scriptsize ±3.10} & 91.74{\scriptsize ±1.32} & \textbf{95.90{\scriptsize ±1.15}} & 73.61{\scriptsize ±5.39} & 64.01{\scriptsize ±7.36} & 88.67{\scriptsize ±0.68} & 75.45{\scriptsize ±2.93} & 65.35{\scriptsize ±0.39} \\ \hline
DIDA & 72.78{\scriptsize ±3.32} & 59.87{\scriptsize ±1.30} & 52.00{\scriptsize ±0.63} & 88.26{\scriptsize ±0.47} & 64.06{\scriptsize ±0.06} & 59.05{\scriptsize ±0.25} & 77.40{\scriptsize ±0.46} & 60.94{\scriptsize ±1.10} & 49.47{\scriptsize ±0.35} & 77.17{\scriptsize ±1.08} & 76.09{\scriptsize ±2.87} & 65.92{\scriptsize ±0.41} \\
Ada-DyGNN & 83.20{\scriptsize ±4.74} & 74.03{\scriptsize ±1.21} & 71.22{\scriptsize ±0.51} & 95.42{\scriptsize ±0.17} & 94.29{\scriptsize ±0.54} & 90.79{\scriptsize ±0.55} & 91.54{\scriptsize ±0.23} & 72.70{\scriptsize ±0.61} & 67.57{\scriptsize ±0.44} & 89.59{\scriptsize ±0.50} & 74.26{\scriptsize ±3.43} & 67.15{\scriptsize ±1.58} \\
EAGLE & 83.41{\scriptsize ±0.97} & 73.12{\scriptsize ±0.36} & 64.46{\scriptsize ±0.37} & 85.07{\scriptsize ±0.86} & 81.71{\scriptsize ±0.44} & 67.38{\scriptsize ±0.35} & 81.51{\scriptsize ±0.56} & 72.41{\scriptsize ±0.31} & 60.69{\scriptsize ±0.59} & 77.90{\scriptsize ±0.11} & 72.19{\scriptsize ±0.61} & 65.06{\scriptsize ±0.55} \\
\textbf{CIR} & \textbf{97.72}{\scriptsize \textbf{±0.13}} & \textbf{90.18}{\scriptsize \textbf{±2.17}} & \textbf{77.46}{\scriptsize \textbf{±4.07}} & \textbf{98.91}{\scriptsize \textbf{±0.06}} & \textbf{98.18}{\scriptsize \textbf{±0.32}} & \textbf{93.82}{\scriptsize \textbf{±1.25}} & \textbf{93.26}{\scriptsize \textbf{±0.71}} & \textbf{80.37}{\scriptsize \textbf{±2.02}} & \textbf{76.04}{\scriptsize \textbf{±4.58}} & \textbf{91.84}{\scriptsize \textbf{±0.44}} & \textbf{76.84}{\scriptsize \textbf{±0.97}} & \textbf{74.88}{\scriptsize \textbf{±0.65}} \\ \hline
\end{tabular}}
\caption{Comparison with SOTA graph link prediction models w.r.t. AUC (mean $\pm$ standard deviation). The best scores are highlighted in bold.}
\label{tab:ood}
\end{table*}

\begin{table*}[t]
\centering
\resizebox{\linewidth}{!}{
\begin{tabular}{c|ccc|ccc|ccc|ccc}
\hline
\textbf{Model} & \multicolumn{3}{c}{\textbf{MOOC}} & \multicolumn{3}{c}{\textbf{REDDIT}} & \multicolumn{3}{c}{\textbf{WIKI}} & \multicolumn{3}{c}{\textbf{UCI}} \\ \hline
Ratio & 0.2 & 0.4 & 0.6 & 0.2 & 0.4 & 0.6 & 0.2 & 0.4 & 0.6 & 0.2 & 0.4 & 0.6 \\ \hline
TGN & 91.51\scriptsize{±1.5} & 59.69\scriptsize{±0.8} & 33.75\scriptsize{±3.2} & 96.47\scriptsize{±0.6} & 86.68\scriptsize{±4.0} & 55.41\scriptsize{±0.5} & 76.18\scriptsize{±1.8} & 64.09\scriptsize{±3.8} & 50.30\scriptsize{±2.8} & 66.82\scriptsize{±2.2} & 64.34\scriptsize{±1.4} & 43.01\scriptsize{±3.2} \\
TGAT & 93.42\scriptsize{±0.3} & 79.92\scriptsize{±1.8} & 62.02\scriptsize{±0.6} & 97.60\scriptsize{±0.1} & 94.64\scriptsize{±0.4} & 81.44\scriptsize{±0.9} & 91.90\scriptsize{±1.5} & 73.06\scriptsize{±0.6} & 59.73\scriptsize{±0.7} & 65.54\scriptsize{±1.0} & 60.84\scriptsize{±0.5} & 52.95\scriptsize{±0.4} \\
GraphMixer & \textbf{98.48\scriptsize{±0.2}} & 87.71\scriptsize{±2.2} & 31.19\scriptsize{±4.9} & \textbf{98.87\scriptsize{±0.3}} & 82.84\scriptsize{±2.6} & 90.85\scriptsize{±2.1} & \textbf{97.13\scriptsize{±0.5}} & 75.68\scriptsize{±4.5} & 61.14\scriptsize{±11.7} & 86.99\scriptsize{±1.5} & \textbf{73.13\scriptsize{±1.4}} & 56.50\scriptsize{±0.6} \\ \hline
DIDA & 67.15\scriptsize{±0.6} & 53.86\scriptsize{±1.2} & 50.74\scriptsize{±0.4} & 81.94\scriptsize{±1.6} & 55.62\scriptsize{±0.2} & 53.97\scriptsize{±0.3} & 73.23\scriptsize{±0.5} & 56.37\scriptsize{±1.0} & 48.07\scriptsize{±0.5} & 69.66\scriptsize{±3.2} & 61.47\scriptsize{±0.8} & 55.38\scriptsize{±0.8} \\
Ada-DyGNN & 81.81\scriptsize{±5.0} & 73.44\scriptsize{±1.4} & 70.04\scriptsize{±1.2} & 94.26\scriptsize{±0.4} & 90.96\scriptsize{±0.4} & 90.51\scriptsize{±0.5} & 92.63\scriptsize{±0.4} & 70.83\scriptsize{±0.3} & 64.54\scriptsize{±0.3} & 88.20\scriptsize{±0.8} & 72.91\scriptsize{±1.8} & \textbf{62.10\scriptsize{±0.8}} \\
EAGLE & 72.10\scriptsize{±0.3} & 72.03\scriptsize{±0.6} & 63.03\scriptsize{±0.6} & 79.82\scriptsize{±0.3} & 77.47\scriptsize{±0.5} & 65.16\scriptsize{±0.5} & 76.01\scriptsize{±0.5} & 66.43\scriptsize{±1.0} & 58.14\scriptsize{±0.5} & 74.06\scriptsize{±0.1} & 62.36\scriptsize{±0.4} & 55.22\scriptsize{±0.4} \\
\textbf{CIR} & \textbf{98.14\scriptsize{±0.5}} & \textbf{88.24\scriptsize{±0.3}} & \textbf{64.53\scriptsize{±2.6}} & \textbf{98.95\scriptsize{±0.1}} & \textbf{98.15\scriptsize{±0.3}} & \textbf{93.65\scriptsize{±2.0}} & 94.96\scriptsize{±0.7} & \textbf{79.10\scriptsize{±4.5}} & \textbf{63.46\scriptsize{±6.0}} & \textbf{90.59\scriptsize{±0.6}} & \textbf{73.55\scriptsize{±1.8}} & 57.36\scriptsize{±0.4} \\ \hline
\end{tabular}
}
\caption{Comparison with SOTA graph link prediction models w.r.t. AP. }
\label{tab:oodap}
\end{table*}

\begin{table*}[t]
\small

\begin{tabular}{l|cccccccc}
\hline
\textbf{Dataset}             & \multicolumn{2}{c}{\textbf{MOOC}}             & \multicolumn{2}{c}{\textbf{REDDIT}}           & \multicolumn{2}{c}{\textbf{WIKI}}             & \multicolumn{2}{c}{\textbf{UCI}}                             \\
                             & AP                    & AUC                   & AP                    & AUC                   & AP                    & AUC                   & AP                    & AUC                   \\ \hline
remove structural encoder & 78.34{\scriptsize±2.50} & 75.12{\scriptsize±0.45} & 89.45{\scriptsize±0.52} & 90.31{\scriptsize±0.38} & 65.82{\scriptsize±2.95} & 63.14{\scriptsize±5.20} & 60.27{\scriptsize±1.68} & 58.91{\scriptsize±2.75} \\
remove temporal encoder   & 80.12{\scriptsize±2.42} & 77.85{\scriptsize±0.41} & 91.62{\scriptsize±0.49} & 92.78{\scriptsize±0.36} & 68.30{\scriptsize±2.88} & 66.42{\scriptsize±5.08} & 63.50{\scriptsize±1.62} & 61.75{\scriptsize±2.60} \\
remove $Y^I$                  & 85.21{\scriptsize±2.30} & 82.64{\scriptsize±0.38} & 95.14{\scriptsize±0.46} & 96.07{\scriptsize±0.33} & 74.45{\scriptsize±2.75} & 72.10{\scriptsize±4.90} & 70.12{\scriptsize±1.55} & 68.35{\scriptsize±2.42} \\
remove $Y^C$  & 81.73{\scriptsize±1.26} & 83.47{\scriptsize±0.22} & 89.31{\scriptsize±0.98} &90.67{\scriptsize±0.52} &70.98{\scriptsize±1.63} &68.49{\scriptsize±1.90} &67.82{\scriptsize±0.96} &62.46{\scriptsize±0.87}\\

CIR                          & \textbf{90.18{\scriptsize±2.17}} & \textbf{88.24{\scriptsize±0.27}} & \textbf{98.18{\scriptsize±0.32}} & \textbf{98.15{\scriptsize±0.25}} & \textbf{80.37{\scriptsize±2.02}} & \textbf{79.10{\scriptsize±4.53}} & \textbf{76.84{\scriptsize±0.97}} & \textbf{73.55{\scriptsize±1.81}} \\ \hline
\end{tabular}
\caption{Ablation study.}
\label{tab:ablation}
\end{table*}

\section{Experiments}
\label{sec:experiments}

We conduct extensive experiments on four benchmark continuous‐time dynamic‐graph datasets to answer the following research questions:
\begin{itemize}[leftmargin=*]
  \item \textbf{RQ1:} How well does CIR mitigate distribution shift?
  \item \textbf{RQ2:} How efficient is CIR compared to existing baselines?
  \item \textbf{RQ3:} What is the contribution of each component?
\end{itemize}
All the experiments are conducted on a computer with Intel(R) Core(TM)2 Duo CPU @2.40 GHz processor, 128 GB RAM, and Tesla T4.

\subsection{Experimental Settings}
\smallskip\noindent\textbf{Datasets}
We conducted experiments on four real-world datasets: \texttt{Wikipedia}\footnote{http://snap.stanford.edu/jodie/wikipedia.csv},
\texttt{Reddit}\footnote{http://snap.stanford.edu/jodie/reddit.csv}
\texttt{MOOC}\footnote{http://snap.stanford.edu/jodie/mooc.csv},
\texttt{UCI}\footnote{https://zenodo.org/records/7213796\#.Y1cO6y8r30o}.  Following prior work~\cite{DIDA}, we introduce structural perturbations to the graphs to simulate distributional shifts. 

\smallskip\noindent\textbf{Evaluation Metrics.}
We partitioned the datasets based on the edge occurrence time: the initial 70\% of edges were designated as the training set, the subsequent 15\% were allocated to the validation set, and the remaining 15\% formed the test set.
We employ the average precision (AP) and area under the curve (AUC) as the evaluation metrics for link prediction. AP and AUC are two common metrics used to evaluate the performance of binary classification models. AP is a measure of the average precision across all possible recall thresholds. AUC is a measure of the area under the receiver operating characteristic (ROC) curve.

\smallskip\noindent\textbf{Training Protocols.}
An early-stopping mechanism was employed, terminating training when the Average Precision (AP) metric showed no improvement for five consecutive epochs. The model underwent training for 300 epochs using the Adam optimizer with a learning rate set at 0.0001 and a weight decay of 1e-6.
 We set the batch size to 600, and the hidden layer dimension to 100.
 For the extraction of the causal subgraph, we specified the number of recent edges ($N$) as 50 and employed 1-hop neighbors. All MLP layers were configured to 2.
Regarding the link prediction task, negative samples were set at a ratio of 1:5 in the training set and adjusted to 1:1 in both the validation and test sets.

\smallskip\noindent\textbf{Baselines.}
Given the limited research on invariant GNNs for CTDGs, 
we consider three categories of baselines:
(1) \textbf{CTDG GNN Models}:
TGN \cite{TGN}, TGAT \cite{TGAT}, and GraphMixer \cite{GraphMixer}. These models are specifically designed for CTDGs but lack the ability to produce explainable and invariant results. Therefore, our comparison primarily focuses on link prediction.
(2) \textbf{Robust DTDG models}:
DIDA \cite{DIDA}, EAGLE \cite{EAGLE}, and Ada-DyGNN \cite{AdaDyGNN}.

\subsection{Link Prediction results (RQ1)}
Table~\ref{tab:ood} reports the AUC results
under varying shortcut ratios (0.2, 0.4, 0.6), while Table~\ref{tab:oodap} presents the corresponding AP scores. Higher values indicate better performance. We summarize the key observations below:
(1) CTDG models degrade sharply under stronger shortcuts.
TGN suffers a substantial performance drop, with AUC decreasing by 31.4\% on MOOC and by 23.8\% on REDDIT.
GraphMixer achieves competitive performance at 0.2 with CIR, but suffering drastic declines, particularly on MOOC (AUC decrease of 50.3\%) and WIKI (AUC decrease of 33.3\%).
TGAT demonstrates relatively better stability but still degrades at higher shortcut ratios.
(2) Robust DTDG models perform relatively well but do not generalize as effectively as CIR.
Since DTDG models operate on discrete time steps, their effectiveness on CTDGs may be constrained, as they are not explicitly designed to handle fine-grained temporal dependencies.
The relatively competitive performance of Ada-DyGNN and EAGLE suggests that their robustness mechanisms still provide advantages, but their full potential might be underestimated due to the CTDG evaluation setting.
(3) CIR achieves the best AUC score across all datasets and shortcut ratios.
Even at the highest shortcut ratio (0.6), CIR maintains strong performance, significantly outperforming other models, demonstrating superior generalization under distributional shifts.
(4) CIR maintains low variance across datasets and shortcut ratios, indicating consistent performance. In contrast, CTDG models, particularly GraphMixer and TGN, exhibit higher variance at larger shortcut ratios, suggesting instability when shortcut dependencies are removed. TGAT shows moderate variance but still struggles with higher shift levels. Among DTDG models, Ada-DyGNN demonstrates lower variance than CTDG models, reflecting its robustness in link prediction, while EAGLE’s variance remains moderate, indicating some inconsistency in handling shifts.

In conclusion, the results demonstrate that CIR achieves SOTA performance in link prediction across varying levels of distribution shift. Unlike existing CTDG and DTDG models, which either suffer from severe performance degradation or exhibit low AUC and AP scores, CIR consistently identifies stable causal structures and maintains high predictive performance with low variance. These results highlight CIR’s strong generalization ability and robustness to spurious correlations in CTDGs.


\begin{figure}[h]
    \centering
    \begin{subfigure}{0.75\linewidth}
\includegraphics[width=\linewidth]{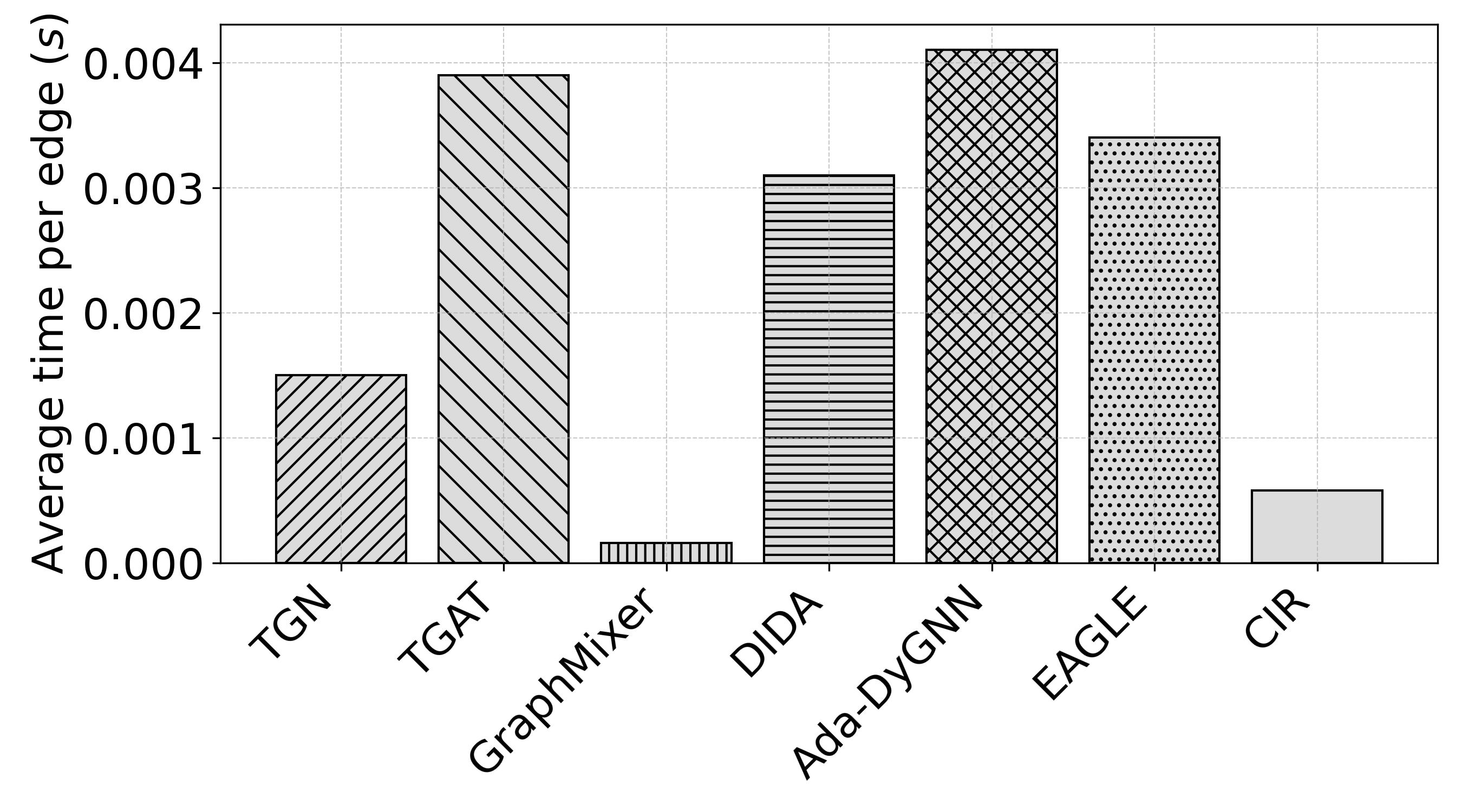}
        \caption{Mooc}
    \end{subfigure}
     \hfill 
    \begin{subfigure}{0.75\linewidth}  \includegraphics[width=\linewidth]{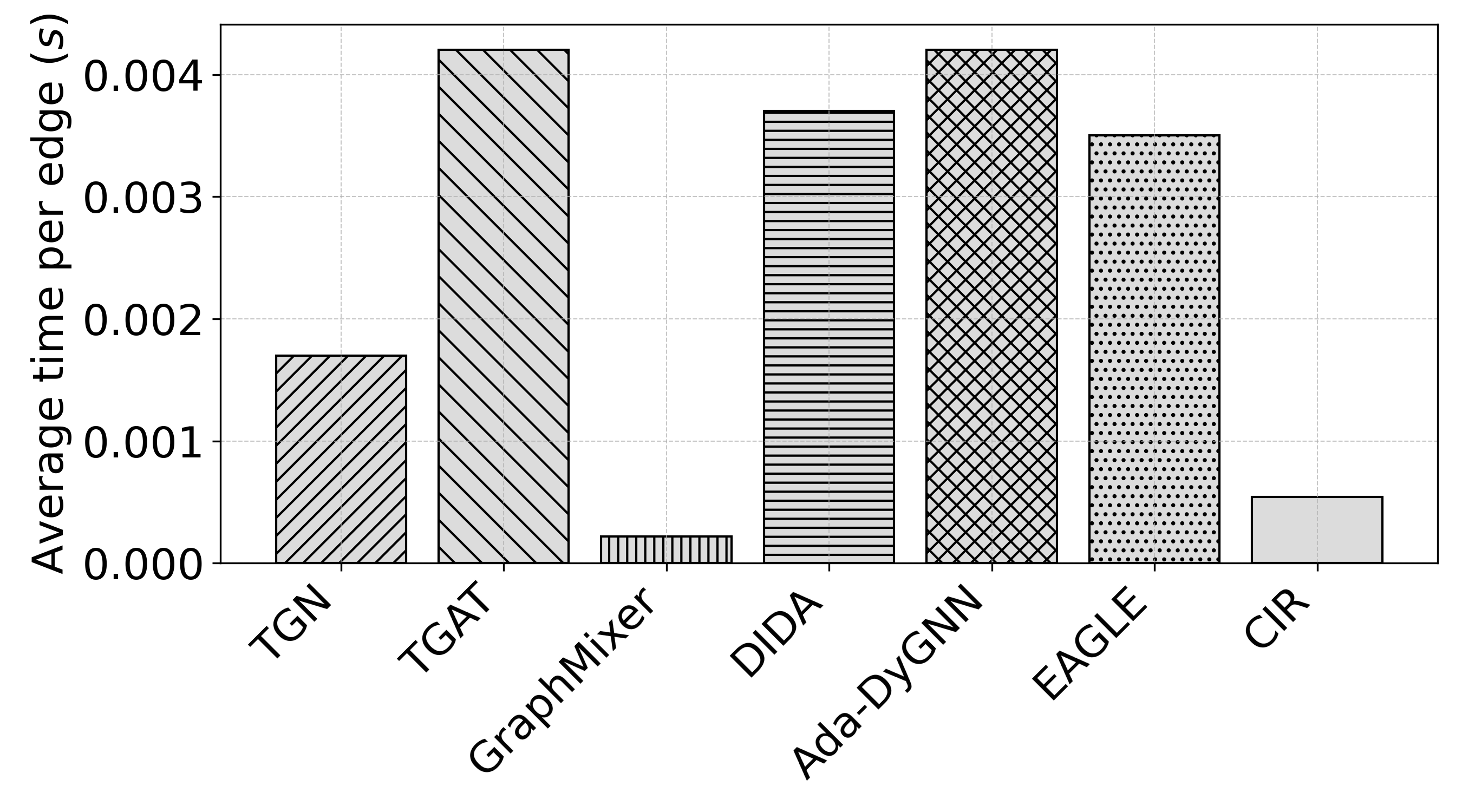}
        \caption{Reddit}
    \end{subfigure}
    \caption{Average running time per edge.}
    \label{fig:time}
\end{figure}

\subsection{Efficiency (RQ2)}
Figure \ref{fig:time} reports the average inference time per edge for each method.
 CIR demonstrates high efficiency, with average running times per edge of  
$5.4\times 10^{-4}$ seconds on Reddit and $5.8\times 10^{-4}$
  seconds on MOOC. In contrast, TGAT and TGN are slower due to their more complex encoding networks, while DIDA is also slower because it must construct a confounder dictionary for each snapshot. GraphMixer is slightly more efficient than CIR, primarily because it does not output an explainable subgraph.
Despite some differences in absolute running times between the Reddit and MOOC datasets, the relative performance across models is consistent, indicating that these efficiency measurements are reproducible across diverse dynamic graph scenarios.

\subsection{Ablation Study  (RQ3)}
Table~\ref{tab:ablation} depicts the ablation study by reporting the AP and AUC across datasets with a fixed shortcut ratio of 0.4. 
The results reveal that each component contributes significantly to the overall performance. 
Removing either the structural encoder or the temporal encoder leads to substantial performance drops, particularly on the WIKI and UCI datasets. 
Moreover, excluding the $Y^C$  causes a marked decline in performance. 
In contrast, the removal of the $Y^I$ has a more modest impact, with performance decreases ranging from about 2\% to 8.9\%. This suggests that while the ICM plays a supportive role in enhancing performance, the CCM is critical for capturing the essential patterns in the data.
In summary, the full CIR model, which integrates all these components, consistently outperforms the ablated versions, highlighting the necessity of each module in achieving optimal performance across diverse datasets.


\section{Conclusion}
This paper addresses the challenge of developing invariant and self-interpretable GNNs for continuous-time dynamic graphs. By analyzing this problem from a causal effect perspective, we introduce the ICCM, a novel causal inference model meticulously designed to address both IID and OOD scenarios for CTDGs. Building upon the theoretical foundations of ICCM, we propose a novel deep learning architecture, which translates theoretically established causal models into a practical solution for dynamic graphs.
Experiments show that CIR outperforms prior methods in link prediction, and robustness under OOD scenarios.

\newpage

\section*{Acknowledgments}
This work was supported by the National Natural Science Foundation of China (No.62472091, 62394333, 62572361)

\bibliographystyle{named}
\bibliography{sample-base}
\end{document}